\documentclass[runningheads]{llncs}

\usepackage[T1]{fontenc}
\usepackage{xcolor}
\usepackage{graphicx}
\usepackage[most]{tcolorbox}
\usepackage{booktabs}
\usepackage{tabularx}
\usepackage[table]{xcolor}
\usepackage{booktabs}
\usepackage{tabularx}
\usepackage[table]{xcolor}
\usepackage{booktabs}
\usepackage[table]{xcolor}

\definecolor{TableHeader}{RGB}{230,242,255}   
\definecolor{RaptorPlus}{RGB}{236,248,238}    

\usepackage{array}
\usepackage{booktabs}
\usepackage{tabularx}
\usepackage[table]{xcolor}
\usepackage{array}
\newcolumntype{Y}{>{\raggedright\arraybackslash}X}

\definecolor{TableHeader}{RGB}{230,242,255} 
\definecolor{CommHdr}{RGB}{255,242,230}     
\definecolor{ZeroHdr}{RGB}{236,248,238}     
\definecolor{FtHdr}{RGB}{240,240,240}       

\definecolor{TableHeader}{RGB}{230,242,255} 
\definecolor{RowHighlight}{RGB}{245,245,245} 
\newcolumntype{Y}{>{\raggedright\arraybackslash}X}

\definecolor{TableHeader}{RGB}{230,242,255} 
\newcolumntype{Y}{>{\raggedright\arraybackslash}X}

\usepackage[dvipsnames,table]{xcolor}

\usepackage{enumitem}

\usepackage{booktabs}     
\usepackage{multirow}     
\usepackage{makecell}     
\usepackage{adjustbox}    
\usepackage{tabularray}   
\UseTblrLibrary{booktabs} 
\usepackage{subcaption}
\usepackage{enumitem}
\usepackage{pifont}       

\usepackage{xspace}
\newcommand{\iou}{IoU\xspace}
\newcommand{\iop}{IoP\xspace}

\newcommand{\raptor}{RAPTOR+\xspace}

\usepackage{amsmath,amssymb,amsfonts,bm,mathtools}
\usepackage{algorithm, algpseudocode} 

\usepackage[section]{placeins} 
\usepackage{rotating}          
\usepackage{pdflscape}         

\usepackage{url}

\usepackage{hyperref}
\usepackage{textcomp}  
\usepackage[english]{babel}
\usepackage{fancyvrb}  

\definecolor{LightCyan}{rgb}{0.88,1,1}
\definecolor{lightskyblue}{RGB}{225, 235, 240}
\definecolor{Gray}{gray}{0.90}
\definecolor{Lightgreen}{RGB}{218, 246, 230}
\usepackage[T1]{fontenc}
\usepackage{graphicx,verbatim}
\usepackage{color}

\begin{document}

\title{RAPTOR+: A Visually Grounded Vision-Language Framework to Improve Clinical Trust and Auditability in Automated Cancer Referral Processing}

\author{Sofiat Abioye\inst{1} \and
Ufaq Khan\inst{2} \and
Shazad Ashraf\inst{3} \and
Anusha Jose\inst{4} \and
Adam Byfield\inst{4} \and
Lukman Akanbi\inst{1} \and
Muhammad Bilal\inst{1,*}} 

\authorrunning{S. Abioye et al.}

\institute{Birmingham City University, Birmingham, England, UK \\
\email{muhammad.bilal@bcu.ac.uk} \and 
Mohamed bin Zayed University of Artificial Intelligence (MBZUAI), Abu Dhabi, UAE \and
University Hospitals Birmingham NHS Foundation Trust, Birmingham, England, UK \and
NHS England, London, England}

\maketitle


\begin{abstract}
Urgent suspected colorectal cancer (CRC) referrals create operational bottlenecks because semi-structured clinical documents often require manual review and transcription. The original RAPTOR system used Large Language Models for structured extraction but relied on a separate OCR stage, making it vulnerable to handwriting, layout variation, and loss of visual evidence linkage.We present \raptor, a multimodal extension that uses Vision-Language Models (VLMs) for end-to-end referral understanding. We evaluate fine-tuned VLMs, commercial and open-source zero-shot VLMs, and the original OCR-based pipeline on 223 clinically curated CRC urgent referral forms. We also introduce a grounding-aware evaluation framework that measures both extraction accuracy and evidence localisation. Results show a clear grounding gap in zero-shot models. Gemini~2.5 Flash achieved 92.6\% Reading Accuracy but only 1.2\% Strict Safety. In contrast, fine-tuned Qwen3-VL-8B achieved 96.1\% Reading Accuracy and 60.6\% Strict Safety, substantially improving verifiable evidence grounding.These findings show that task-specific fine-tuning is essential for reliable, auditable clinical document understanding. \raptor enables extracted referral decisions to be linked to visual evidence, supporting safer and more efficient cancer referral triage.
\end{abstract}


\section{Background}
\label{sec:background}

Reducing delays in the urgent diagnostic pathway for suspected cancer is a major health-system priority~\cite{graber2025interventions}. In many healthcare
systems, key referral information is exchanged between clinicians as semi-structured documents: scanned forms, faxed documents, and handwritten annotations that require manual review and transcription into downstream systems. This creates substantial administrative burden and
introduces the risk of omission or transcription error, potentially delaying cancer diagnoses~\cite{roberts2025characterising}.

Urgent suspected colorectal cancer (CRC) referral pathways exemplify these challenges. CRC referrals combine free-text clinical narratives with structured elements such as checkboxes and tables, and frequently include handwritten or low-contrast annotations~\cite{laing2024codesigned}. Poor
scan quality, administrative stamps, strikethroughs, overwritten entries, and heterogeneous templates across referring practices can undermine OCR-first pipelines, which depend on clean text extraction before any downstream information extraction can occur~\cite{chen2022automated,abioye2025raptor}.

Recent vision-language and document foundation models provide an alternative, producing structured outputs directly from document images using joint representations of text, layout, and visual features. Performance on generic document benchmarks has been strong, but this
does not guarantee safety or reliability in real-world clinical deployment contexts~\cite{huang2022layoutlmv3,kim2022ocr,khan2026medobvious}.

Two failure modes are particularly problematic in urgent referral processing. \emph{Ungrounded extractions} occur when models generate plausible values not supported by the source document~\cite{ji2023survey}. \emph{Non-verifiable predictions} arise when an extracted value may be
correct, but the system cannot indicate where the supporting evidence appears, preventing clinician validation~\cite{chen2022towards}. Both failure modes undermine trust and limit the practical utility of large-scale
automated extraction systems in clinical workflows where accountability and auditability are paramount~\cite{sharma2024advances,gill2021hybrid,rasheed2025federated}.

\begin{figure}[H]
  \centering
  \includegraphics[width=\linewidth]{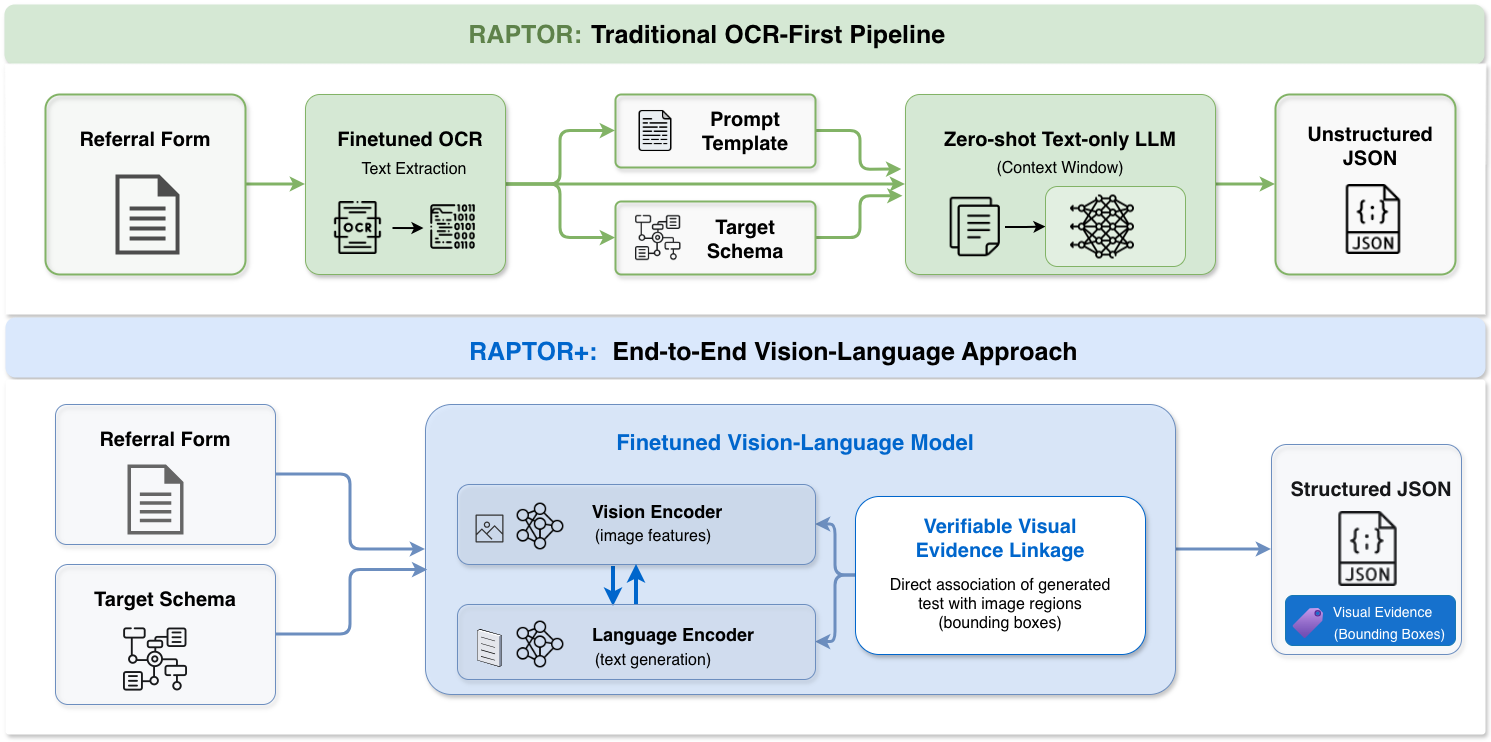}
  \caption{\textbf{Overview of \raptor vs.\ RAPTOR.}
    \textit{Top panel:} RAPTOR follows an OCR-first pipeline: the form
    is processed via OCR/text extraction and combined with the expected
    schema as input to a zero-shot text-only LLM, producing JSON outputs
    without reliable visual evidence linkage.
    \textit{Bottom panel:} \raptor performs end-to-end, visually
    grounded extraction by conditioning a fine-tuned vision-language
    model on the referral form and the expected schema to produce
    structured JSON outputs.}
  \label{fig:overview}
\end{figure}

We introduce \raptor (Figure~\ref{fig:overview}), a grounding-aware evaluation framework for end-to-end vision-language understanding of urgent CRC referral forms. It jointly measures structured extraction accuracy and evidence localisation quality. We benchmark representative
vision-language models, characterise clinically relevant failure modes, and analyse how model scale and adaptation affect grounding reliability for human-in-the-loop deployment.

The contributions of this work are threefold. First, we propose a grounding-aware evaluation framework for urgent CRC referrals that jointly quantifies structured extraction accuracy and visual evidence localisation. Second, we benchmark representative end-to-end document
VLMs and characterise clinically relevant failure modes, including ungrounded extractions and poor evidence localisation quality. Third, we analyse the effects of model scale and task-specific adaptation on
grounding reliability and discuss implications for safe deployment in urgent referral triage workflows~\cite{miao2025minimum}.

\section{Methods}
\label{sec:methods}

\subsection{Dataset and Experimental Setup}

We conducted experiments on a synthetic, clinically curated dataset of 223 urgent CRC referral documents.
The dataset reflects real-world variability, including heterogeneous layout structures, free-text entries, incomplete fields, handwritten annotations, and editorial marks such as strikethroughs and overwritten
entries. All referral images were standardised to $1654 \times 2339$~pixels, with bounding-box coordinates normalised for cross-model comparison. The dataset was partitioned into 176 training forms and a disjoint
held-out test set of 47 forms, with no overlap in document templates or patient scenarios. Ground-truth supervision included structured target values for each clinical field, field-presence logic indicating whether each field should contain a value or remain null, and precise bounding-box annotations identifying the visual evidence supporting each extracted value.

\subsection{Models and Training Regimes}

We evaluated a diverse set of VLMs spanning commercial API-access
systems and open-weight alternatives, to isolate how model variants,
scale, and training regime affect structured extraction with geometric
evidence (Table~\ref{tab:models}).

Commercial models evaluated in zero-shot mode included Claude Opus~4.6
and Gemini~2.5 Flash, evaluated in pure inference mode without
parameter updates or task-specific training. Open-weight models
evaluated in zero-shot mode included the Qwen3-VL variants (2B, 4B, 8B,
and 32B parameters), allowing analysis of how scale influences grounding
behaviour without task-specific training~\cite{liu2023large}. All zero-shot
models used identical prompting strategies. Open-weight models evaluated
after fine-tuning included Qwen3-VL variants (2B, 4B, and 8B
parameters), trained on paired document images and structured JSON
targets containing field values and bounding-box coordinates for
supporting evidence~\cite{bai2025qwen3,wu2025selfai}. Fine-tuning used standard
supervised learning with cross-entropy loss on value prediction and
coordinate regression.

All models were prompted with identical task specifications to emit a
single JSON object adhering to a fixed schema, with one evidence
bounding box per field. Bounding boxes were produced in each model's
native coordinate convention and normalised for evaluation, so
performance differences reflected model behaviour rather than prompt
engineering.

\begin{table}[H]
  \centering
  \caption{Vision-language models evaluated in this study. Fine-tuned
    models use the same architecture as their base counterparts,
    differing only in the training regime.}
  \label{tab:models}
  \small
  \begin{tabularx}{\linewidth}{@{}lXccc@{}}
    \toprule
    \textbf{Model} & \textbf{Source} & \textbf{Params}
      & \textbf{Regime} & \textbf{Role in Study} \\
    \midrule
    \multicolumn{5}{@{}l}{\textit{Commercial (closed-source)}} \\
    \addlinespace[2pt]
    Gemini 2.5 Flash  & Closed source & N/A & Zero-shot & SOTA commercial baseline \\
    Claude Opus 4.6   & Closed source & N/A & Zero-shot & SOTA commercial baseline \\
    \addlinespace[4pt]
    \multicolumn{5}{@{}l}{\textit{Open-weight (zero-shot)}} \\
    \addlinespace[2pt]
    Qwen3-VL-32B      & Open source & 32B & Zero-shot & Large-scale open baseline \\
    Qwen3-VL-8B       & Open source &  8B & Zero-shot & Mid-scale base model \\
    Qwen3-VL-4B       & Open source &  4B & Zero-shot & Scaling baseline \\
    Qwen3-VL-2B       & Open source &  2B & Zero-shot & Scaling baseline \\
    \addlinespace[4pt]
    \multicolumn{5}{@{}l}{\textit{Open-weight (fine-tuned)}} \\
    \addlinespace[2pt]
    Qwen3-VL-8B~(FT)  & Open source &  8B & Fine-tuned & Primary intervention \\
    Qwen3-VL-4B~(FT)  & Open source &  4B & Fine-tuned & Scaling baseline \\
    Qwen3-VL-2B~(FT)  & Open source &  2B & Fine-tuned & Scaling baseline \\
    \bottomrule
  \end{tabularx}
\end{table}

\subsection{System Architecture}

\begin{figure}[H]
  \centering
  \includegraphics[width=0.9\linewidth]{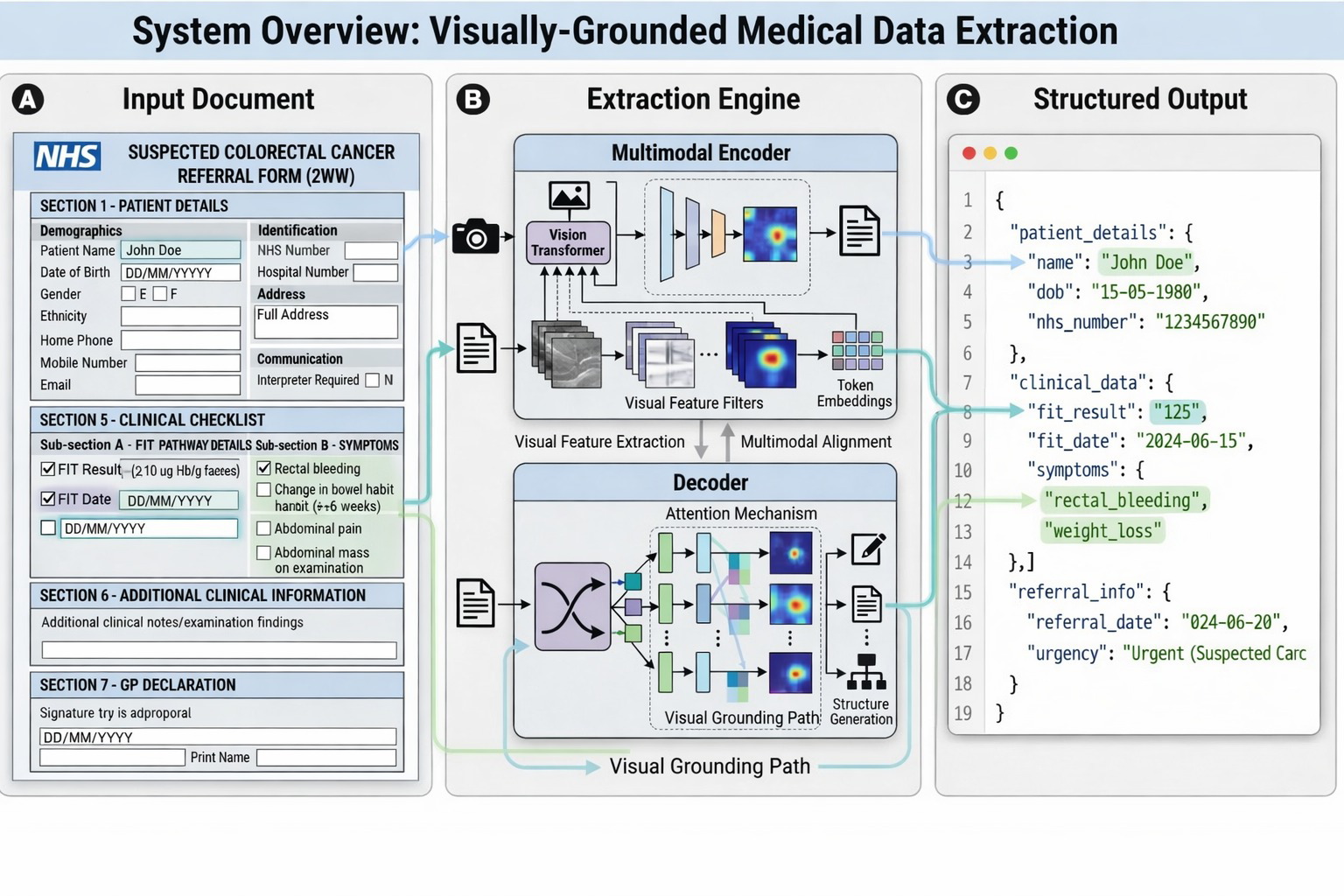}
  \caption{\textbf{System overview of the \raptor visually grounded medical data extraction pipeline.} The architecture consists of three core stages \textbf{(A)}~\textit{Input Document}, featuring an urgent CRC referral form with targeted clinical fields;
    \textbf{(B)}~\textit{Extraction Engine}, a multimodal
    transformer-based encoder-decoder that processes visual features via a Vision Transformer and aligns them with linguistic tokens; and \textbf{(C)}~\textit{Structured Output}, which generates grounded clinical data in JSON format. The \textit{Visual Grounding Paths} (coloured arrows) demonstrate the model's ability to provide verifiable evidence by tracing extracted values such as patient
    identifiers, FIT results, and clinical symptoms, directly back to their source bounding boxes on the raw document scan.}
  \label{fig:architecture}
\end{figure}

The \raptor architecture has three processing stages
(Figure~\ref{fig:architecture}). The \textit{input} stage accepts an
urgent CRC referral form containing clinical fields such as patient
demographics, symptoms, examination findings, and diagnostic test
results. The \textit{extraction engine} is a multimodal
transformer-based encoder-decoder that processes visual features through
a Vision Transformer and aligns them with linguistic tokens via
cross-attention. The \textit{output} stage produces grounded clinical
data in structured JSON format, with each field value accompanied by
bounding-box coordinates for the supporting visual evidence.

Visual grounding paths within the architecture enable verifiable
evidence by tracing extracted values directly back to their source
regions on the raw document scan~\cite{vasey2022reporting}. This end-to-end
design removes the decoupled OCR stage of the original RAPTOR system.
The model learns joint representations of visual appearance, layout, and
semantic content, and the grounded output allows clinicians to verify
each extraction against the source document.

\subsection{Evaluation Protocol}

We evaluated performance at the field level on the held-out test set of
47 forms using a multi-criteria protocol that separates semantic
extraction quality from evidence localisation quality.

\textbf{Value Correctness (Reading Accuracy / NM).} For each field, we
compared the predicted value to the ground-truth value after
domain-aware normalisation resolving case differences, whitespace and
punctuation variations, numeric and unit formatting inconsistencies, and
common Boolean or empty-value synonyms. The primary reading metric was
\emph{Normalised Match} (NM), which canonicalised surface forms to focus
evaluation on semantic content rather than formatting artefacts.

\textbf{Evidence Reliability (Evidence Precision / EP).} We quantified
the propensity to produce ungrounded evidence by computing Evidence
Precision~\cite{bai2025qwen3}, defined as the proportion of predicted
bounding boxes that overlapped with any ground-truth evidence region.
Low EP indicates frequent hallucinated pointing, where models generate
geometrically invalid bounding boxes not corresponding to actual
document regions.

\textbf{Strict Safety (Automated Compliance, IoU\,$\geq$\,0.5).} We
evaluated whether predicted evidence boxes enabled fully automated
verification by computing Intersection-over-Union (\iou) between
predicted and ground-truth evidence boxes
~\cite{tong2023rethinking}. A
prediction was counted as compliant if \iou\,$>$\,0.5, a conservative
criterion penalising shifted, overly tight, or excessively loose
bounding boxes. The \emph{Joint} metric requires both correct value
extraction and a compliant evidence box simultaneously.

\textbf{Audit Precision (Human-in-the-Loop Pointing, IoP\,$\geq$\,0.5).}
We captured cases where models indicated the correct region without
precisely matching the annotated extent using
Intersection-over-Prediction (\iop)~\cite{tong2022ngiou}. A
prediction was considered useful for clinical audit if
\iop\,$>$\,0.5, aligning with clinician review behaviour where a
reasonably tight box in the correct region can substantially accelerate
verification.

\section{Results}
\label{sec:results}

\subsection{Semantic Value Extraction Performance}

Zero-shot and commercial models demonstrated strong semantic extraction
capabilities. Claude Opus~4.6 achieved 95.2\% reading accuracy, and
Qwen3-VL-32B reached 95.3\%, the highest among all zero-shot models.
Gemini~2.5 Flash achieved 92.6\%. These results indicated that
large-scale pre-training on diverse document corpora enabled robust
handling of heterogeneous templates, mixed free-text and tabular
regions, and handwritten annotations characteristic of real-world
clinical referral forms.

Fine-tuning improved both semantic extraction accuracy and grounding
reliability. Qwen3-VL-8B improved from 88.4\% reading accuracy in
zero-shot mode to 96.1\% after fine-tuning, surpassing all zero-shot
models while also gaining substantially in grounding quality. The
fine-tuned model learned to align value extraction with accurate
evidence bounding boxes, reflecting a fundamental shift in model
behaviour induced by task-specific supervision.

The original RAPTOR system, which employed GPT-4o-mini and DeepSeek-R1
with decoupled OCR, achieved the highest semantic accuracy at 94.83\%
and 93.72\% respectively~\cite{abioye2025raptor}. However, these systems
provided no localisation capability for audit purposes, limiting their
utility in safety-critical deployment scenarios requiring human
verification.

Table~\ref{tab:results} summarises all quantitative results.

\begin{table}[H]
  \centering
  \caption{Quantitative results. Reading accuracy (value match) and
    grounding metrics. \textbf{Strict Safety} requires correct value
    extraction \emph{and} a compliant evidence box (\iou\,$>$\,0.5).
    \textbf{Pointing Precision} reports audit-oriented localisation
    success (\iop\,$>$\,0.5). `---' denotes models without
    bounding-box output. Bold = best per column; italics = second
    best.}
  \label{tab:results}
  \small
  \setlength{\tabcolsep}{7pt}
  \begin{tabular}{@{}llccc@{}}
    \toprule
    \multirow{2}{*}{\textbf{Model}}
      & \multirow{2}{*}{\textbf{Regime}}
      & \textbf{Reading Acc.}
      & \textbf{Strict Safety}
      & \textbf{Pointing Prec.} \\
      & & \textit{(Value Match)}
      & \textit{(Joint \iou)}
      & \textit{(Joint \iop)} \\
    \midrule
    \multicolumn{5}{@{}l}{\textit{Commercial (closed-source)}} \\
    \addlinespace[2pt]
    Gemini 2.5 Flash  & Zero-shot  & 92.6\%          & 1.2\%  & 41.4\% \\
    Claude Opus 4.6   & Zero-shot  & 95.2\%          & 0.2\%  &  1.0\% \\
    \addlinespace[6pt]
    \multicolumn{5}{@{}l}{\textit{Open-weight (zero-shot)}} \\
    \addlinespace[2pt]
    Qwen3-VL-32B      & Zero-shot  & \textit{95.3\%} & 8.0\%  & \textit{60.9\%} \\
    Qwen3-VL-8B       & Zero-shot  & 88.4\%          & 6.2\%  & 37.2\% \\
    Qwen3-VL-4B       & Zero-shot  & 87.5\%          & 0.3\%  & 33.3\% \\
    Qwen3-VL-2B       & Zero-shot  & 62.3\%          & 0.0\%  & ---    \\
    \addlinespace[6pt]
    \multicolumn{5}{@{}l}{\textit{Open-weight (fine-tuned)}} \\
    \addlinespace[2pt]
    Qwen3-VL-8B~(FT)  & Fine-tuned & \textbf{96.1\%} & \textbf{60.6\%} & \textbf{75.4\%} \\
    Qwen3-VL-4B~(FT)  & Fine-tuned & 90.1\%          & \textit{25.9\%} & 46.4\% \\
    Qwen3-VL-2B~(FT)  & Fine-tuned & 88.3\%          & 21.0\%          & 32.9\% \\
    \bottomrule
  \end{tabular}
\end{table}

\subsection{Evidence Presence and Hallucination Behaviour}

\begin{figure}[H]
  \centering
  \includegraphics[width=1.0\linewidth]{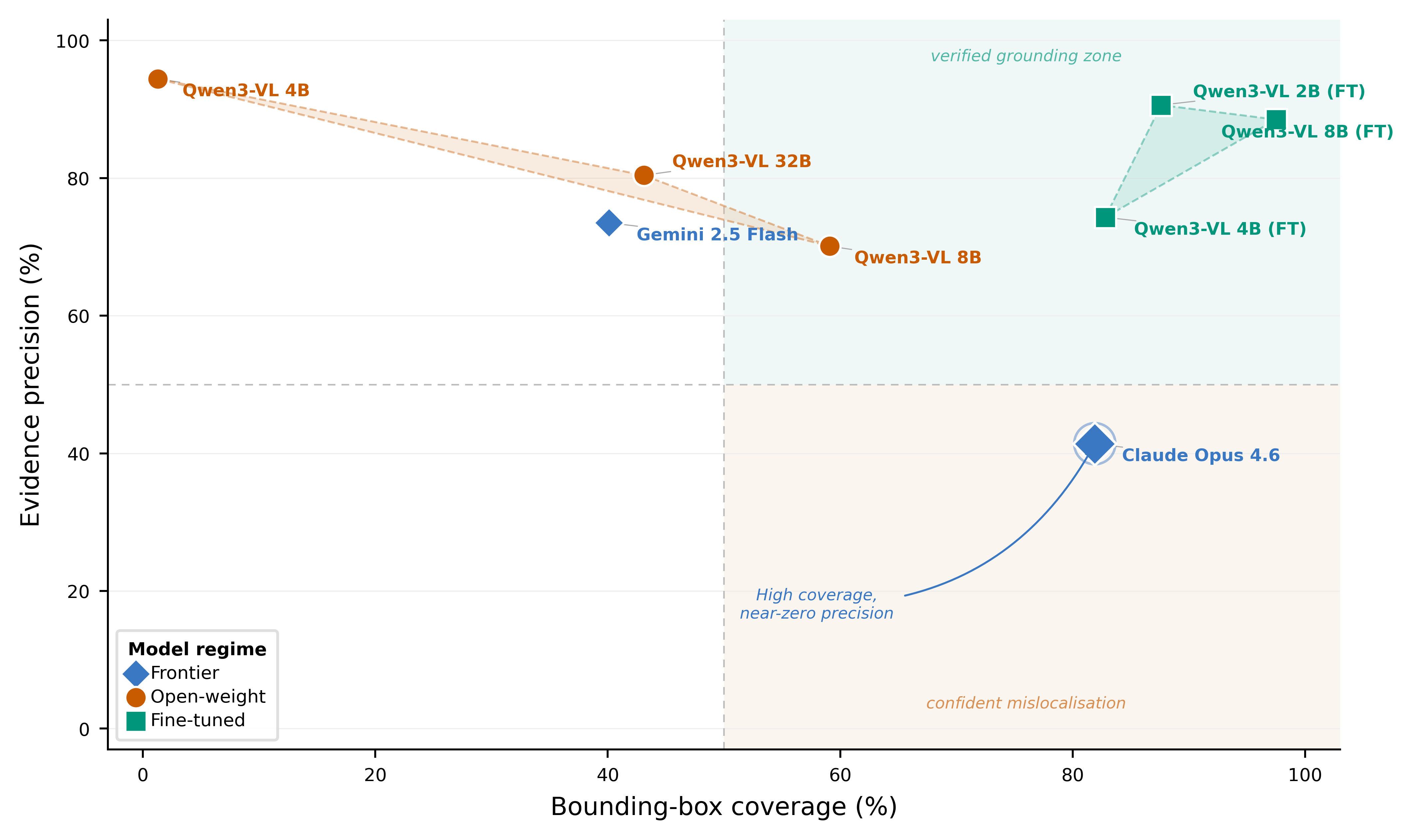}
  \caption{\textbf{Evidence Precision vs.\ Bounding-Box Coverage.}
    Each point represents one model; colour and shape denote regime
    (blue diamond = Frontier; orange circle = Open-weight; green
    square = Fine-tuned). The dashed lines at 50\% mark the
    high-performance zone. Fine-tuning moves models towards high
    coverage \emph{and} high precision, bridging the grounding gap.}
  \label{fig:ep_scatter}
\end{figure}

Zero-shot models varied considerably in evidence precision.
Gemini~2.5 Flash achieved 73.5\% EP, while Qwen3-VL-32B reached 80.4\%.
Claude Opus~4.6, despite its high reading accuracy, achieved only 41.4\%
EP, indicating that its bounding boxes frequently did not correspond to
actual evidence regions. These results show that strong reading accuracy
does not guarantee reliable grounding, a critical gap that undermines
trust and prevents reliable automated verification
(Figure~\ref{fig:ep_scatter}).

Claude Opus~4.6 represents a particularly important failure mode:
confident mislocalisation. Unlike the smaller open-weight models that
simply withheld bounding boxes, Claude Opus produced coordinates for
81.9\% of fields, generating the appearance of thorough grounding while
achieving only 0.2\% Strict Safety and 1.0\% Audit Precision. The model
was prolific in its pointing yet almost entirely wrong. From a clinical
safety perspective this is worse than silence: a clinician auditing
extractions against highlighted evidence regions would be directed to
incorrect parts of the document on virtually every field, undermining
rather than supporting verification. High bounding-box coverage paired
with near-zero \iou or \iop is therefore a distinct and dangerous
failure mode that aggregate reading accuracy metrics cannot detect.

Fine-tuning further improved evidence precision. Qwen3-VL-8B after
fine-tuning achieved 88.5\% EP, up from 70.1\% in zero-shot mode. The
fine-tuned model generated bounding boxes more reliably, substantially
reducing ungrounded evidence generation. This improvement, combined with
gains in reading accuracy, represents a highly favourable outcome for
safety-critical applications.

\subsection{Joint Task Success: Value and Grounding}

Joint task success, defined as simultaneously extracting the correct
value and producing a verifiable evidence box, revealed stark
differences between zero-shot and fine-tuned models. Under strict
safety criteria requiring \iou\,$\geq$\,0.5, zero-shot models showed
limited joint success despite high reading accuracy. Gemini~2.5 Flash
achieved only 1.2\% joint success, while Qwen3-VL-32B reached 8.0\%
and Qwen3-VL-8B 6.2\%. These results demonstrate that semantic
extraction capability alone is insufficient for safety-critical
deployment.

\begin{figure}[H]
  \centering
  \includegraphics[width=\linewidth]{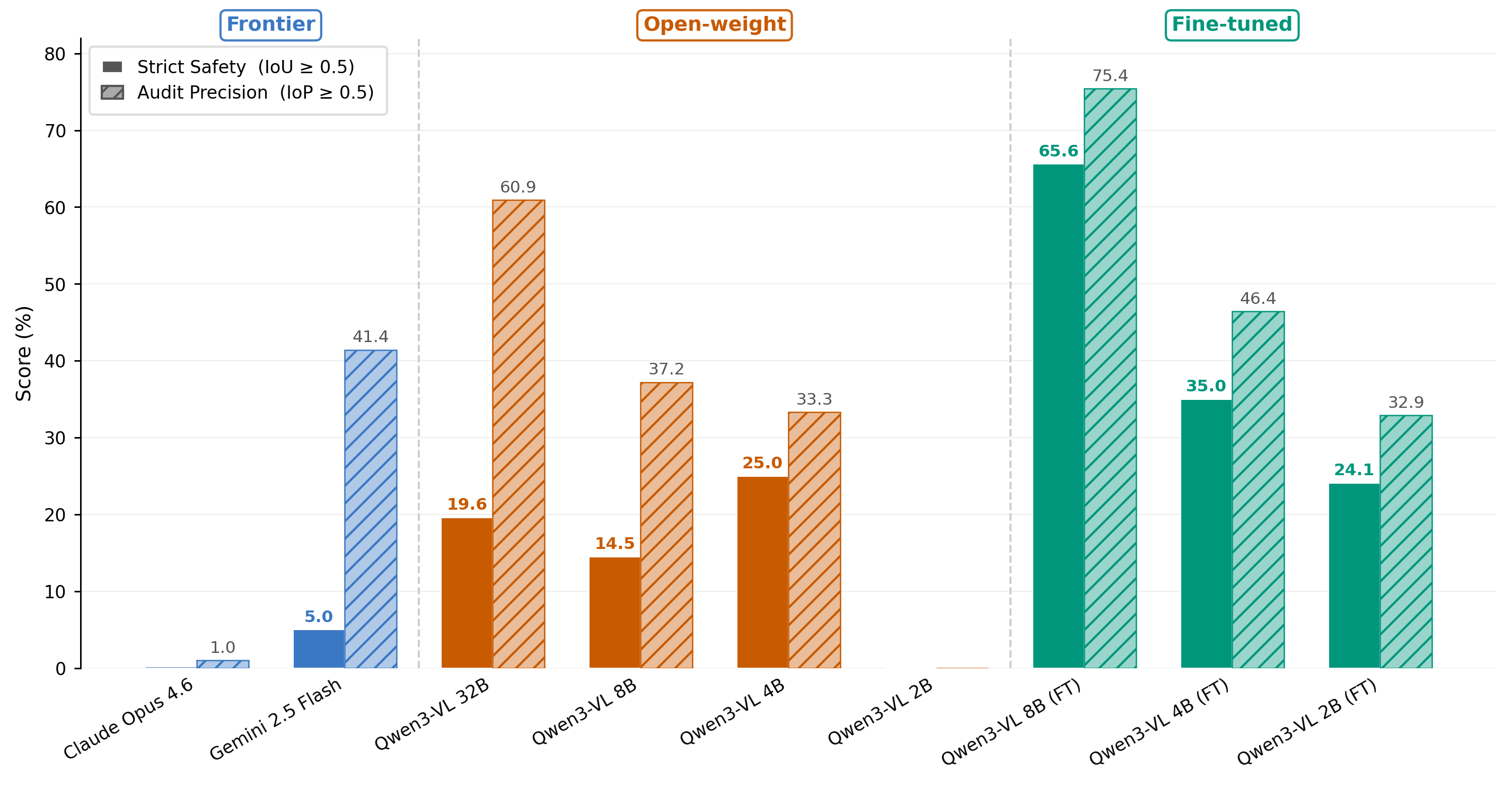}
  \caption{\textbf{Spatial localisation metrics across model tiers.}
    Paired bars show Strict Safety (\iou\,$\geq$\,0.5, darker) and
    Audit Precision (\iop\,$\geq$\,0.5, lighter) per model, grouped by
    regime. Fine-tuned models substantially outperform zero-shot and
    commercial baselines on both metrics.}
  \label{fig:spatial}
\end{figure}

In contrast, Qwen3-VL-8B after fine-tuning reached 60.6\% strict joint
success, validating task-specific fine-tuning as necessary for
geometric guarantees~\cite{nunes2024health}. The spatial precision advantage
is visualised in Figure~\ref{fig:spatial}. This represents approximately
a tenfold improvement over the Qwen3-VL-8B zero-shot baseline,
confirming that supervised learning on paired value-evidence examples
enabled reliable grounding.

Under the more permissive audit precision criterion requiring
\iop\,$>$\,0.5, large zero-shot models demonstrated partial
recovery~\cite{salwei2024usability}. Qwen3-VL-32B achieved 60.9\% joint
success under \iop, while Gemini~2.5 Flash reached 41.4\%. These
results indicate that zero-shot models can correctly identify values and
point to approximately correct regions for a meaningful proportion of
fields, making them potentially effective as audit assistants in
human-in-the-loop workflows despite their limitations for fully
automated verification.

\subsection{Effect of Fine-Tuning on Grounded Understanding}

Fine-tuning functioned as a structural switch for open-weight VLMs,
converting strong generic reading capability into schema-aligned,
evidence-grounded extraction~\cite{hartsock2024vision}. The base Qwen3-VL-8B
model demonstrated strong reading accuracy of 88.4\% but limited
geometric compliance at 6.2\% strict safety. After task-specific
supervision, the fine-tuned variant achieved 60.6\% strict safety,
representing approximately a tenfold gain. This improvement pattern
indicates that fine-tuning primarily taught the model where evidence
boundaries lie rather than improving text reading capability per se.

Task-specific supervision further improved evidence
precision~\cite{lust2022efficient}. EP increased from 70.1\% for the base
model to 88.5\% after fine-tuning, suggesting the model learned to
produce more precise coordinate predictions aligned with actual evidence
regions. Fine-tuning improved both reading accuracy
(88.4\%\,$\to$\,96.1\%) and grounding quality simultaneously, showing
no trade-off between extraction quality and evidence reliability. In
clinical audit settings, a high-recall system with reliable evidence
pointers enables rapid verification and maintains trust.

For smaller open-weight models, fine-tuning acted not as a refinement
of existing grounding capability but as its activation from scratch. In
zero-shot mode, Qwen3-VL-4B produced bounding boxes for only 1.3\% of
fields, and Qwen3-VL-2B produced none at all, rendering their
localisation output effectively absent. The 94.4\% evidence precision
reported for Qwen3-VL-4B zero-shot is therefore misleading in
isolation: it reflects precision on the rare occasions the model fired,
not a reliable grounding signal across the form. After fine-tuning,
both models entered functional grounding territory, with bounding-box
coverage rising to 82.8\% and 87.6\% respectively. This demonstrates
that task-specific supervision unlocked the localisation mechanism
rather than merely calibrating it. For sub-8B deployment in
resource-constrained clinical settings, fine-tuning is consequently a
prerequisite for any grounding capability, not an optional enhancement.

\subsection{Trade-off Analysis: Accuracy Versus Auditability}

The \raptor evaluation revealed the cost of neglecting grounding in prior systems. The original RAPTOR system prioritised semantic accuracy, achieving 94.83\% reading accuracy but providing no localisation for audit. \raptor with fine-tuned Qwen3-VL-8B achieved
both high semantic accuracy (96.1\%) and grounded, verifiable evidence, achieving 60.6\% to 75.4\% visual auditability depending on the strictness criterion applied.

This result reflects a principled design choice for safety-critical applications~\cite{gyevnar2025ai}. In urgent cancer referral triage, the ability to rapidly verify extraction decisions and trace them to source evidence is essential for maintaining clinician trust and ensuring accountability. A system with slightly lower recall but reliable evidence pointers enables efficient human oversight, whereas a high-accuracy system without grounding capability requires complete manual review, negating automation benefits.

\section{Conclusion}
\label{sec:conclusion}

\raptor jointly measures structured extraction accuracy and evidence localisation, revealing failure modes that text-only metrics obscure. Large zero-shot models read clinical documents well but show limited strict grounding: Gemini~2.5 Flash reached 92.6\% semantic accuracy yet
only 1.2\% Strict Safety (Joint \iou).Task-specific fine-tuning closed this gap. The fine-tuned Qwen3-VL-8B
achieved 60.6\% strict joint success and 88.5\% evidence precision, a 50-fold improvement over the best zero-shot commercial baseline (1.2\%). Fine-tuned models frequently output \texttt{null} when localisation is uncertain, accepting lower recall in exchange for reliable pointers. In a human-in-the-loop workflow, these abstentions are flagged for clinician review rather than propagating unverified values downstream. Zero-shot models, though currently unsuitable for fully automated verification, retain utility in human-in-the-loop workflows, directing clinician attention to relevant regions for a meaningful proportion of fields. These results support evidence-linked, human-in-the-loop deployment in urgent referral processing. Future work should examine generalisation to other referral types, robustness under real-world distribution shift, and clinician qualitative evaluation of grounding quality in operational workflows.

\bibliography{main}

\end{document}